\documentclass[fleqn,10pt]{wlscirep}
\usepackage[utf8]{inputenc}
\usepackage[T1]{fontenc}
\usepackage{multirow}
\usepackage{subfig}

\title{Study on the identification limits of craniofacial superimposition}

\author[1,2,*]{Oscar Ibáñez}
\author[3]{Enrique Bermejo}
\author[2]{Andrea Valsecchi}
\affil[1]{Faculty of Computer Science, CITIC, University of A Coruña, A Coruña 15071, Spain	}
\affil[2]{Panacea Cooperative Research S. Coop., Ponferrada, Spain}
\affil[3]{Andalusian Research Institute in Artificial Intelligence (DaSCI), University of Granada, Spain}
\affil[*]{oscar.ibanez@udc.es}

\begin{abstract}
 Craniofacial Superimposition involves the superimposition of an image of a skull with a number of ante-mortem face images of an individual and the analysis of their morphological correspondence. Despite being used for one century, it is not yet a mature and fully accepted technique due to the absence of solid scientific approaches, significant reliability studies, and international standards. In this paper we present a comprehensive experimentation on the limitations of Craniofacial Superimposition as a forensic identification technique. The study involves different experiments over more than 1 Million comparisons performed by a landmark-based automatic 3D/2D superimposition method. The total sample analyzed consists of 320 subjects and 29 craniofacial landmarks.
\end{abstract}
\begin{document}

\flushbottom
\maketitle
%
%
\thispagestyle{empty}

\section*{Introduction}
HUMAN IDENTIFICATION (ID) of the dead is of paramount importance in our society. It does not only resolve serious legal and social predicaments (prove the death of a family member), but also it provides a resolution to grieving families who need closure to their sadness. The most accepted methods of identification (DNA; fingerprints; dental) are often inappropriate or even impossible when there is not enough information available, ante-mortem (AM) or post-mortem (PM), due to the state of preservation of the corpse or the lack of data. While the skeleton usually survives both natural and non-natural decomposition processes (fire, salt, water, etc.), the soft tissue progressively degrades and is lost, as is the most frequent case in disaster victim identification (DVI). Water/fire-damage, decomposition and delayed access to the bodies may confound any comparative methods. The socioeconomic conditions of the countries of origin and the resulting lack of regular dental treatment, fingerprint resources and family records make directly impossible to apply them in many scenarios. In fact, the experience of several practitioners indicates the poorer effectiveness of DNA analysis (around 3\% of the IDs) and dactyloscopy (15-25\%) against skeleton-based identification (SFI) techniques (70-80\%) in DVI scenarios.

Craniofacial Identification (CI) \cite{stephan_carl_n_claes_peter_craniofacial_2016} methods represent an alternative for ID using techniques such as Forensic Facial Comparison (FFC) \cite{bacci_forensic_2021}, Craniofacial Reconstruction \cite{wilkinson_facial_2010} or Comparative Radiography (CR), Craniofacial Superimposition (CFS) \cite{damas_handbook_2020}. Among them, CFS is the only comparative method that can be applied in cases of badly perserved facial soft tissue.

CFS is probably the most challenging skeleton-based identification method. It involves the superimposition of an image of a skull with a number of ante-mortem face images of an individual and the analysis of their morphological correspondence. This skull-face overlay (SFO) process is usually done by corresponding anatomical (anthropometric) landmarks located in the skull (craniometric) and the face (cephalometric). Thus, two objects of different nature are compared (a face and a skull). CFS fundamentals rely on the uniqueness of both the face and the skull as well as their strong relationship, since the skull is the main factor determining the shape of the face.

The potential for the application of CFS is enormous. The only AM data required is a facial photograph. The democratization of smartphones and the increasing use of social networks makes it easier to have access to a large number of recent facial photographs worldwide. The required PM material is the skull. CFS can be always applied, as accurate images of the bony tissue can be obtained using Magnetic Resonance (MRI), Computer Tomography (CT) and Cone Bean CT (CBCT) scans or simpler devices such as light structure scans or photogrammetry systems when there is direct access to the skull. 
 
The objective of the present study is to assess the limitations of craniofacial superimposition (CFS) as a forensic identification technique as well as to analyze the relationship between craniometric and cephalometric anatomical landmarks and facial soft tissue thickness (FSTT).

\section*{Results}

The experiments performed can be differentiated in two groups according the landmark set considered, and the number of total comparisons. The first group of results is compiled in Table \ref{tab:1} and corresponds to three experiments (E1-E3); the most comprehensive experiment on the power of CFS as identification technique for the number of comparisons involved. The results of six different experiments are shown, each of them involving more than 1 million CFS cross-comparisons (321 Skulls VS 3210 synthetic facial photos) and 29 landmarks located in each pair of skull and face images.

\begin{table}[]
\centering
\begin{tabular}{ccccccccc}
\textbf{Exp} & \textbf{\# Land  } & \textbf{\# Sbj } & \textbf{\# SFO } & \textbf{FSTT} & \textbf{ST Direction} & \textbf{Noise } & \textbf{Rank (avg)} & \textbf{Accuracy (\%)} \\ \hline
E1                            & 29                                  & 320                               & 1024000                           & Real                           & Real                                   & 0                                & 1                                    & 100                                     \\ \hline
E2                            & 29                                  & 320                               & 1024000                           & None                           & Real                                   & 0                                & 1                                    & 100                                     \\ \hline
\multirow{4}{*}{E3}           & \multirow{4}{*}{29}                 & \multirow{4}{*}{320}              & \multirow{4}{*}{1024000}          & \multirow{4}{*}{Mean}          & \multirow{2}{*}{Real}                  & 0                                & 1.69                                 & 90                                      \\
                              &                                     &                                   &                                   &                                &                                        & 5 px                             & 2.55                                 & 82.1                                    \\ \cline{6-9} 
                              &                                     &                                   &                                   &                                & \multirow{2}{*}{Mean}                  & 0                                & 6.39                                 & 74.8                                    \\
                              &                                     &                                   &                                   &                                &                                        & 5 px                             & 8.16                                 & 66.6                                    \\ \hline
\end{tabular}
\caption{Summary of the results for E1-E3. Legend of the performed experiments: (E1) complete knowledge regarding soft-tissue thickness (FSTT = real) and spatial (ST Direction = real), absence of landmark location errors (Noise = 0); (E2) no soft-tissue is modeled, as this experiment involves a the superimposition of two skulls (3D/2D); (E3) mean populational  soft-tissue thickness (FSTT = mean) and spatial (ST Direction = real/mean) data. For each sub-experiment, absence of landmark location errors (Noise = 0) 5x5 pixels error in landmark location (Noise = 5).\label{tab:1}}

\end{table}

\begin{figure}[ht]
\centering
\includegraphics[width=\linewidth]{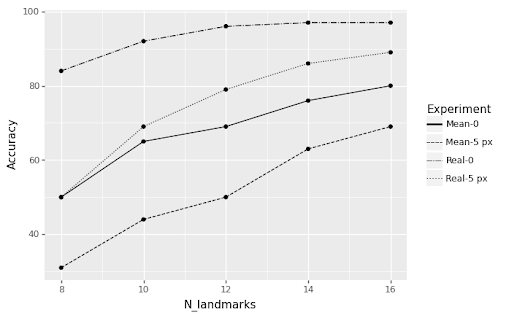}
\caption{Comparison of the performance for the CFS scenarios designed in E4.
 Accuracy by number of visible landmarks and FSTT direction vector.}
\label{fig:1}
\end{figure}

Table \ref{tab:2} shows the results of E4, performed over a different set of landmarks and less subjects, although involving different combinations of number of landmarks, direction of the soft-tissue vector and localization noise to simulate the effect of bias in human observers. The results of 20 different experiments are shown, each of them involving 49K CFS cross-comparisons (70 Skulls VS 700 synthetic facial photos) and 29 landmarks located in each pair of skull and face images. A random set of visible landmarks from the total is selected for each si Figure \ref{fig:1} graphically summarizes the results of E4 in terms of accuracy.

\begin{table}[]\centering

\begin{tabular}{llllllllll}
\textbf{Exp       }           & \textbf{\#Land }              & \textbf{\#Sbj }               & \textbf{\#SFO}                   & \textbf{FSTT}                         & \textbf{\#Vis }               & \textbf{Dir}                   & \textbf{Noise} & \textbf{Averaged Rank} &\textbf{ Acc (\%)} \\ \hline
\multirow{20}{*}{E4} & \multirow{20}{*}{29} & \multirow{20}{*}{70} & \multirow{20}{*}{49000} & \multirow{20}{*}{Mean} & \multirow{4}{*}{8L}  & \multirow{2}{*}{Real} & 0     & 1.80          & 84       \\
                     &                      &                      &                         &                              &                      &                       & 5 px  & 3.92          & 50       \\ \cline{7-10} 
                     &                      &                      &                         &                              &                      & \multirow{2}{*}{Mean} & 0     & 5.47          & 50       \\
                     &                      &                      &                         &                              &                      &                       & 5 px  & 7.96          & 31       \\ \cline{6-10} 
                     &                      &                      &                         &                              & \multirow{4}{*}{10L} & \multirow{2}{*}{Real} & 0     & 1.45          & 92       \\
                     &                      &                      &                         &                              &                      &                       & 5 px  & 2.26          & 69       \\ \cline{7-10} 
                     &                      &                      &                         &                              &                      & \multirow{2}{*}{Mean} & 0     & 4.03          & 65       \\
                     &                      &                      &                         &                              &                      &                       & 5 px  & 6.12          & 44       \\ \cline{6-10} 
                     &                      &                      &                         &                              & \multirow{4}{*}{12L} & \multirow{2}{*}{Real} & 0     & 1.28          & 96       \\
                     &                      &                      &                         &                              &                      &                       & 5 px  & 1.83          & 79       \\ \cline{7-10} 
                     &                      &                      &                         &                              &                      & \multirow{2}{*}{Mean} & 0     & 3.12          & 69       \\
                     &                      &                      &                         &                              &                      &                       & 5 px  & 4.28          & 50       \\ \cline{6-10} 
                     &                      &                      &                         &                              & \multirow{4}{*}{14L} & \multirow{2}{*}{Real} & 0     & 1.30          & 97       \\
                     &                      &                      &                         &                              &                      &                       & 5 px  & 1.57          & 86       \\ \cline{7-10} 
                     &                      &                      &                         &                              &                      & \multirow{2}{*}{Mean} & 0     & 2.41          & 76       \\
                     &                      &                      &                         &                              &                      &                       & 5 px  & 3.5           & 63       \\ \cline{6-10} 
                     &                      &                      &                         &                              & \multirow{4}{*}{16L} & \multirow{2}{*}{Real} & 0     & 1.21          & 97       \\
                     &                      &                      &                         &                              &                      &                       & 5 px  & 1.42          & 89       \\
                     &                      &                      &                         &                              &                      & \multirow{2}{*}{Mean} & 0     & 2.08          & 80       \\
                     &                      &                      &                         &                              &                      &                       & 5 px  & 2.52          & 69       \\ \hline
\end{tabular}
\caption{Summary of the results for E4. All experiments involve a variable amount of visible landmarks (\#Vis), mean populational  soft-tissue thickness (FSTT = mean) and different configurations of spatial (ST Direction = real/mean) data, and noise in the landmark location errors (Noise = 0/5).\label{tab:2}}

\end{table}

\section*{Discussion}

In the last years, the CFS field has experienced significant outcomes. Firstly, the international standardization effort has occurred under the umbrella of the European project MEPROCS \cite{damas_handbook_2020} between 2011 and 2014. Secondly, a number of reliability and validation studies have been carried out showing contradictory results. Finally, the fourth generation of CFS systems has developed. In these systems different AI techniques are employed to automate one or more stages of the CFS process, making its application faster, more accurate, and objective.

The results achieved by both experiments provide invaluable insight on the importance of knowledge extraction processes, the strong correlation between the shape of a human skull and a face, and the solid foundations of CFS as identification procedure.
 
\section*{Methods}

\subsection*{Dataset}
The dataset \cite{guyomarch_facial_2013} comprises 3D coordinates corresponding to anatomical landmarks using a software tool. The sample consists of 500 CT-scans of living French individuals with known age and sex. Sex ratio is 1:1.13 (265 males; 235 females), mean age is 52 years (range = 18–96; standard deviation s = 20), and constitutes a representative sample of the French population. Landmarks were annotated with TIVMI software (Treatment and Increased Vision in Medical Imaging) after a surface reconstruction of both bone and soft tissue. The Frankfurt Horizontal plane was used to ease landmark localization. Different sets of subjects were subsampled according the set of landmarks involved in the experimentation. A preliminary analysis was performed to filter those subjects where landmarks were deemed incorrectly located by the human observers due to large errors in their localization.

\subsection*{Landmarks}
A total number of  86 landmarks (see \ref{tab:3} were manually annotated over the considered dataset, both in the bone and its corresponding point in the soft tissue. That way, the true FSTT is known for each landmark pair. However, due to the procedure used to obtain the CT-scans, some cases were partial scans, resulting in incomplete data. The experimentation was constructed using different subsets of landmarks to maximize the number of cases available and study the robustness of the technique before different landmark sets. 

\begin{table}[h!]
\centering
\begin{tabular}{cll}
\textbf{\#} & \textbf{E1, E2 \& E3 }               & \textbf{E4   }                       \\ \hline
1  & Glabella                    & Glabella                    \\
2  & Left Dacryon                & Gnathion                    \\
3  & Left Ectoconchion           & Left Ectoconchion           \\
4  & Left Frontomalare Orbitale  & Left Ectomolare\_2          \\
5  & Left Frontotemporale        & Left Frontomalare Orbitale  \\
6  & Left Mid-supraorbital       & Left Frontotemporale        \\
7  & Left Nasomaxillare          & Left Mentale                \\
8  & Left Orbitale               & Left Mid-nasomaxillare      \\
9  & Left Superciliare           & Left Mid-ramus              \\
10 & Left Supraorbital Ridge     & Left Nasomaxillare          \\
11 & Left Zygion                 & Left Orbitale               \\
12 & Left Zygomatic              & Left Superciliare           \\
13 & Left Zygomaxillare          & Left Supra Canine           \\
14 & Left Zygoorbitale           & Left Zygion                 \\
15 & Nasion                      & Mid-philtrum                \\
16 & Rhinion                     & Nasion                      \\
17 & Right Dacryon               & Prosthion                   \\
18 & Right Ectoconchion          & Right Ectoconchion          \\
19 & Right Frontomalare Orbitale & Right Ectomolare\_2         \\
20 & Right Frontotemporale       & Right Frontomalare Orbitale \\
21 & Right Mid-supraorbital      & Right Frontotemporale       \\
22 & Right Nasomaxillare         & Right Mentale               \\
23 & Right Orbitale              & Right Mid-nasomaxillare     \\
24 & Right Superciliare          & Right Mid-ramus             \\
25 & Right Supraorbital Ridge    & Right Nasomaxillare         \\
26 & Right Zygion                & Right Orbitale              \\
27 & Right Zygomatic             & Right Superciliare          \\
28 & Right Zygomaxillare         & Right Supra Canine          \\
29 & Right Zygoorbitale          & Right Zygion              \\ \hline
\end{tabular}
\caption{Landmark sets considered for each experiment performed.\label{tab:3}}
\end{table}

\begin{figure}[ht]
\centering
\subfloat[]{\includegraphics[width=0.5\textwidth]{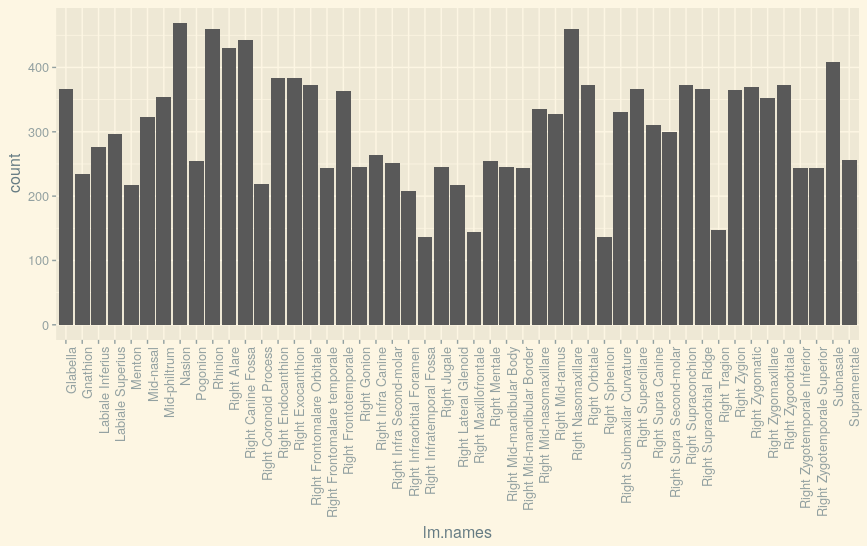}}
  \hfill
  \subfloat[]{\includegraphics[width=0.4\textwidth]{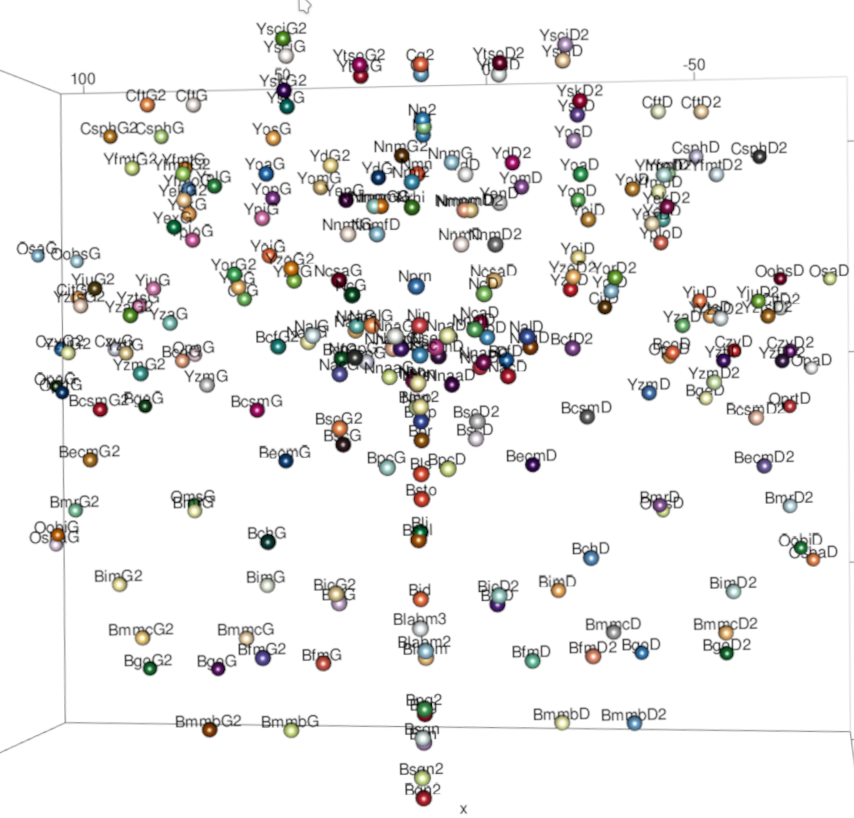}}
\caption{(a) Distribution of the 86 (12 unilateral) landmarks annotated in the dataset. (b) 3D Visualization of landmark coordinates.}
\label{fig:2}
\end{figure}

 \subsection*{Procedure}
From the 3D coordinates provided in the dataset, a series of synthetic photographs were generated to simulate 3D/2D CFS scenarios according to corresponding sets of landmarks. Two poses were considered (frontal and lateral), and five photographs were simulated per subject and pose. In order to simulate the uncertainty of manual localization of landmarks when performed by human observers, an alternative test was considered by randomly displacing landmarks up to 5 pixels in the 2D photographs. The following parameters were used:
\begin{itemize}
    \item Focal range [0.5, 1.5]
    \item Pose rotation [-15º, 15º]
    \item Image resolution: Width [600,1200] | Height [600,1000]
    \item Landmark localization noise [0px, 5px]
\end{itemize}

\begin{figure}[ht]
\centering
\subfloat{\includegraphics[width=0.4\textwidth]{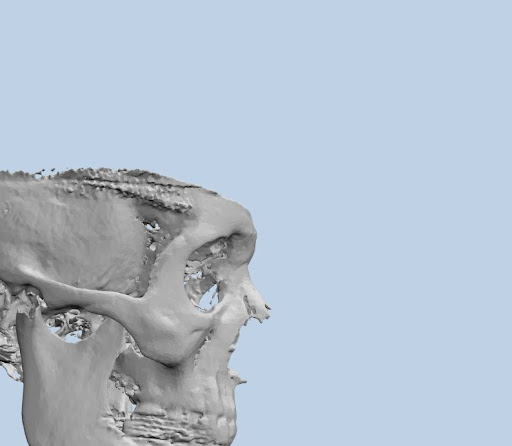}}
  \hfill
  \subfloat{\includegraphics[width=0.46\textwidth]{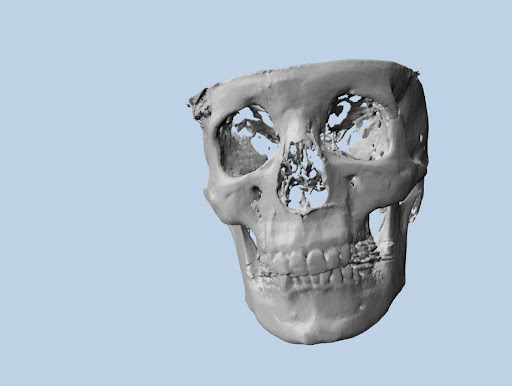}}
\caption{Example of a frontal and lateral synthetic photograph for a skull, showing the impact of randomized simulations.}
\label{fig:3}
\end{figure}

\begin{figure}[ht]
\centering
\subfloat{\includegraphics[width=0.45\textwidth]{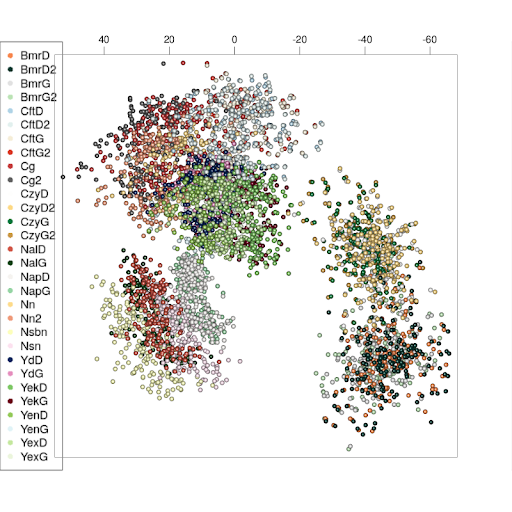}}
  \hfill
  \subfloat{\includegraphics[width=0.45\textwidth]{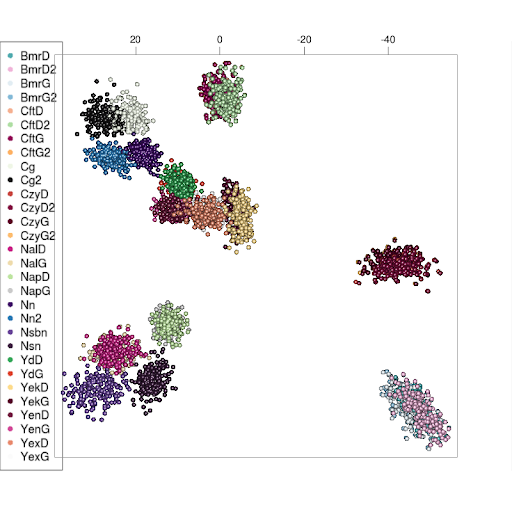}}
\caption{Left, original 3D distribution of landmarks for all the subjects in the dataset. Right, landmarks aligned to a common coordinate system after performing PCA. }
\label{fig:4}
\end{figure}

Prior to generating the simulated photographs, PCA was performed to establish a common coordinate system and align the subjects (see Figure\ref{fig:3}). The POSEST-SFO algorithm \cite{valsecchi_robust_2018} solves a system of polynomial equations relating the distances between the points before and after the projection. It was used to perform the 3D-2D superimposition of each scenario. After overlaying a 3D model of a subject with a photograph, a similarity metric is computed for all the landmarks visible in the photograph. This metric is then used to obtain a ranking across the whole set of synthetic photographs (questioned individuals) to determine the accuracy of the identification. 

\subsection*{Experiment 1 - Human Face Singularity}
This experiment considers a craniometric landmark set with no uncertainty involved, as we consider the real FSTT and its real direction vector. Such a setting is similar to using cephalometric landmarks, and allows us to replicate a landmark-based 3D/2D facial superimposition setting, that is, overlaying a projected 3D face to a facial photograph. Next, the number of landmarks, subjects and identification scenarios or cross-comparisons for E1 is summarized:
\begin{itemize}
\item 29 landmarks (3 unilateral and 12 bilateral)
\item 321 subjects
\item 3210 simulated photographs
\item Average \# of visible landmarks: 18-28  frontal poses | 6-14 lateral poses
\item 1.030.410 NxN comparisons or skull-face overlays (SFO)
\end{itemize}

\subsection*{Experiment 2 - Human Sull Singularity}
This second experiment aims to reproduce the exact same setting as E1, only using craniometric landmarks. That is, the experiment involves a direct comparison among skulls of the different subjects, no soft-tissue is modeled. Instead of simulating facial photographs, 2D photographs of skulls are simulated in different poses according the random parameters considered and 3D skulls are superimposed onto the photographs. The purpose of this experiment is to analyze the individualization power of craniometric landmarks and study the performance of the superimposition approach. 

\subsection*{Experiment 3 - Craniofacial superimposition}
The third experiment consists on overlaying a projection of a 3D skull onto a 2D facial photograph. With this experiment, we study the influence of FSTT by considering an averaged value of the soft-tissue vector from the population sample, simulating the uncertainty of a real identification scenario. To compare with scenarios with different levels of uncertainty, both the real and the averaged direction of this vector are compared.

\subsection*{Experiment 4 - Landmark influence in CFS}
The final experiment considers a different landmark set (see Table~\ref{tab:1}) which covers a comprehensive area of the face. The purpose of this experiment is to analyze the performance of CFS depending on landmark visibility. Next, the number of landmarks, subjects and identification scenarios or cross-comparisons for E4 is summarized:
\begin{itemize}
    \item 29 landmarks (3 unilateral and 12 bilateral)
    \item 70 subjects
    \item 700 simulated photographs
    \item Variable \# landmarks for each experiment in [8-16 landmarks]
    \item 49000 NxN comparisons per experiment
\end{itemize}

A total of five sub-experiments have been carried out by fixing the number of landmarks visible in the facial photographs. Each experiment is subdivided by considering the averaged FSTT of the whole dataset population and the following combinations: i) real direction vector, ii) mean direction vector, iii) no noise, and iv) random noise in landmark localization (2D).

\bibliography{ExportedItems}



\section*{Acknowledgements}

We wish to acknowledge the support received from the Centro de Investigación de Galicia "CITIC", funded by Xunta de Galicia and the European Union (European Regional Development Fund- Galicia 2014-2020 Program), by grant ED431G 2019/01. Oscar Ibáñez was supported by the Ministry of Science under grant RYC2020-029454-I. Dr. Bermejo's work is funded by the Regional Government of Andalusia [Ref: DOC\_01130]. Funding for open access charge: Universidade da Coruña/CISUG. 

Additionally, This work was supported by the Ministry of Science and Innovation [Ref: PID2021-122916NB-I00] and the Ministry of Economic Affairs and Digital Transformation and the entity Red.es [Ref:2021/C005/00141299].

\section*{Author contributions statement}
O.I. conceived the experiment(s),  E.B. and A.V. conducted the experiment(s), E.B. and A.V. analysed the results.  All authors reviewed the manuscript. 

\section*{Additional information}

\textbf{Competing interests} The authors declare that they have no known competing financial interests or personal relationships that could have appeared to influence the work reported.


\end{document}